\RequirePackage[l2tabu, orthodox]{nag}

\documentclass[12pt,msc,a4paper,oneside]{ucl_thesis}

\usepackage{blindtext}

\usepackage{emptypage}

\usepackage{graphicx}

\usepackage{float}

\usepackage{amsmath}

\usepackage{gensymb}
\usepackage{textcomp}

\usepackage{setspace}
\usepackage{multirow}

\usepackage{bibentry} 

\usepackage[format=hang,font=small,labelfont=bf]{caption}

\usepackage{etoolbox}

\usepackage{bibentry}
\makeatletter\let\saved@bibitem\@bibitem\makeatother
\usepackage[pdftex,hidelinks]{hyperref}
\makeatletter\let\@bibitem\saved@bibitem\makeatother
\makeatletter
\AtBeginDocument{
    \hypersetup{
        pdfsubject={Thesis Subject},
        pdfkeywords={Thesis Keywords},
        pdfauthor={Author},
        pdftitle={Title},
    }
}
\makeatother

\title{Character-level Convolutional Network for Text Classification Applied to Chinese Corpus}
\author{Weijie Huang}
\department{Department of Computer Science}

\begin{document}

\nobibliography*


\maketitle
\makedeclaration

\begin{abstract} 
Compared with word-level and sentence-level convolutional neural networks (ConvNets), the character-level ConvNets has a better applicability for misspellings and typos input. Due to this, recent researches for text classification mainly focus on character-level ConvNets. However, while the majority of these researches employ English corpus for the character-level text classification, few researches have been done using Chinese corpus. This research hopes to bridge this gap, exploring character-level ConvNets for Chinese corpus test classification. We have constructed a large-scale Chinese dataset, and the result shows that character-level ConvNets works better on Chinese character dataset than its corresponding pinyin format dataset, which is the general solution in previous researches. This is the first time that character-level ConvNets has been applied to Chinese character dataset for text classification problem.
\end{abstract}
\begin{acknowledgements}
I gratefully acknowledge the support of my family. My mom and dad have provided me with the chance to study aboard, and I applicate it. I would like to thank Supervisor Dr. Jun Wang, Ph.D. student Yixin Wu, Rui Luo for their helpful feedback and advice. I have learned a lot from them during the meeting every week. Their enthusiasm for the research encourages me to finish this thesis. And also I would like to thank Institute of Education student, my roommate, Qing Xiang. She corrects lots of grammar mistakes that I did not realize in her spare time. Finally, I would like to thank Arsene Wenger that brings several new players from transfer market this two months, so that I can mainly focus on my thesis with an optimistic attitude.
\end{acknowledgements}

\setcounter{tocdepth}{2} 

\tableofcontents
\listoffigures
\listoftables

\chapter{Introduction}

\section{Background}

Natural language processing (NLP) is the field in which through analysing data, the machine can extract information from contexts and represent the input information in a different way\cite{conneau2016very}. Generally speaking, NLP involves the following three tasks. Part-Of-Speech tagging (POS), such as text or image classification, to classify the data with different categories; Chunking (CHUNK), to label the segment of a given sentence by using syntactic and semantic relations; and Named Entity Recognition (NER), to tag named entities in text\cite{collobert2011natural}. These tasks are varied, ranging from character-level to word-level and even to sentence-level. Nevertheless, they have the same purpose of finding out the hierarchical representations of the context\cite{conneau2016very}.

One of the classic tasks for NLP is text classification, also known as document classification\cite{collobert2008unified}. This task aims to assign a pre-defined label to the document. Usually, two stages are involved in the process which are feature extraction and labels classification. In the first stage, some particular word combinations such as bigram, trigram, term frequency, and inverse document frequency of the phrase can be used as features.\cite{collobert2011natural} Take BBC sports website for example, in its content, there are many specific Premier League team names, which can serve as corresponding features for the following classification. These features can then, in the second stage, help to maximise the accuracy of the task.

\section{Previous solutions}

A common approach to text classification is to use Bag of Words\cite{harris1954distributional} , N-gram\cite{cavnar1994n}, and their term frequency-inverse document frequency (TF-IDF)\cite{sparck1972statistical} as features, and traditional models such as SVM\cite{joachims1998text}, Naive Bayes\cite{mccallum1998comparison} as classifiers. However, recently, many researchers\cite{zhang2015character}\cite{collobert2011natural}\cite{kim2014convolutional}\cite{conneau2016very}, using deep learning model, particularly the convolutional neural networks (ConvNets), have made significant progress in computer vision\cite{he2015deep} and speech recognition\cite{abdel2012applying}. ConvNets, originally invented by LeCun\cite{lecun1998gradient} for computer vision, refers to the model that uses convolution kernels to extract local features. Analyses have shown that ConvNets is effective for NLP tasks\cite{zhang2015shallow}\cite{severyn2015unitn}, and the convolution filter can be utilised in the feature extraction stages. Compared with the model listed above, ConvNets, when applied to text classification has shown rather competitive results\cite{zhang2015character}\cite{collobert2011natural}\cite{kim2014convolutional}\cite{conneau2016very}. The theory behind is quite similar to that of computer vision and speech recognition task. During the process in the convolutional layers, convolutional kernels would first treat the input text as a 1D-image. Then by using a fixed size convolution kernel, it can extract the most significant word combination such as the “English Premier League” in the sports topics. After hierarchical representations of the context are constructed, these features are then fed into a max-pooling layer for feature extraction, and the output result can represent the most important feature on this topic. In other words, the 1D-ConvNets can be regarded as a high-level N-grams feature classifier. With the help of ConvNets, we can classify the unlabelled document without using syntactic or semantic structures of a particular language. This is unusual in most of the large-scale dataset. Also, since the ConvNets method can handle the misspelling problem, it works well for user-generated data\cite{zhang2015character}.

Recent approach of using ConvNets on text classification mainly works at the word-level \cite{collobert2011natural} \cite{kim2014convolutional}. In Kim\' s research, he found that pre-trained words embedding could gain a slight improvement on performances. Also, the multi-channel model allows the randomly initialised tokens to learn more accurate representations during the task. When different regularisers are applied, dropout layer proved to work well in the task, increasing by 4\% in relative performance. Although this approach has achieved great success, some limitations remain. Firstly, when word-level ConvNets is applied in the classification task, the words sharing a common root, prefix or suffix tend to be treated as separate words. For instance, the words ``surprise'' and ``surprisingly'' are treated as two words without any relation in the model, which is counterintuitive. Secondly, the dimension of the input layer is related to the dictionary of the words. Due to the common root issue, the dimension up to 6,000 may lead to a sparse problem, which will significantly influence the performance. Thirdly, the words which are not presented in the training set will be marked as out-of-vocabulary words (OOV) and then simply replaced with a blank character. This problem frequently occurs in our test corpus, which may lead to serious consequences. Also, since some of the classification datasets are postings directly collected from the social network, these corpora are mixed with typos and abbreviations. This may diminish the classification accuracy of the task\cite{xiao2016efficient}.

In the last few years, many researchers found that it is also likely to train a ConvNets at the character-level\cite{zhang2015character}\cite{xiao2016efficient}\cite{conneau2016very}. The researchers still used one-hot or one-of-m encoding, and the vectors are transformed from the raw character to the dense vectors. Kim used a character sequence as an input in his language model\cite{kim2015character}, and Dhingra applied this idea to predict the hashtags\cite{dhingra2016tweet2vec}. The character-level ConvNets can avoid the problems as mentioned earlier from word-level. Firstly, since the units in the model are now the character, we can prevent the problem that some words sharing the same prefix or suffix do not show any relations. Secondly, the dictionary of character-level ConvNets are the size of the alphabet plus some symbols, and the dimension is around 70 in most of the character-level model. Due to this small dimension, the sparse problem can now be solved. Thirdly, since the choice of the alphabet is the same in both the training set and the test set, no more OOV appears in the test dataset. A significant improvement can be observed in which the model can better handle the typos with fewer parameters.

\begin{figure}[t]
\centering 
\fbox{\includegraphics[width = 9cm]{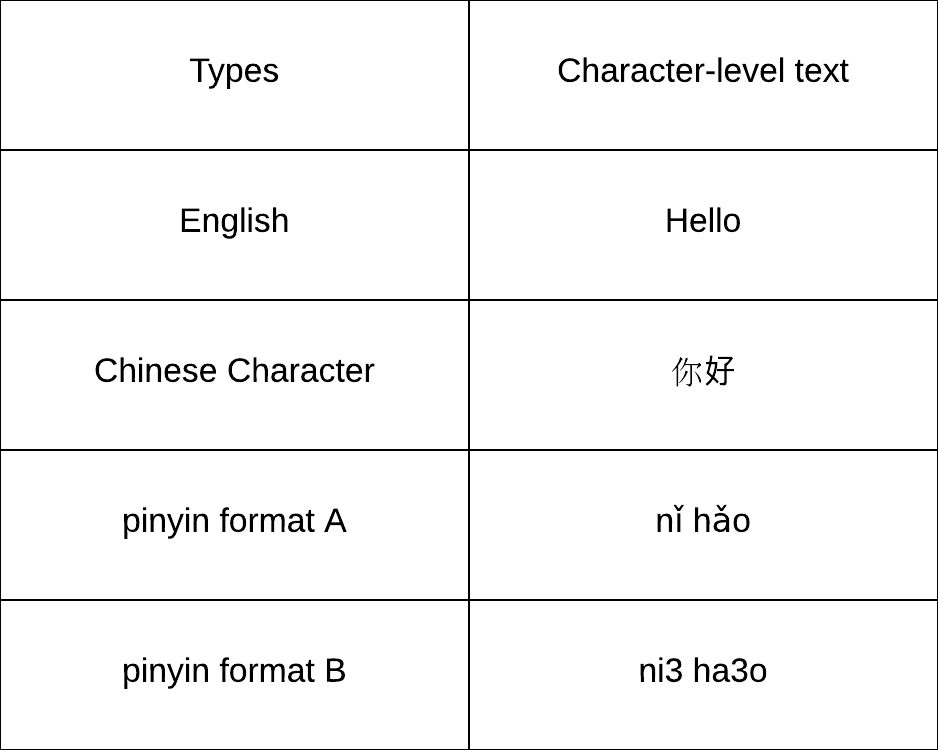}}
\caption{The relation between Chinese character and its pinyin format. Listed in the figure are two types of pinyin encoding format A and B. Type A combines the tone and character, while type B separate the tone and character.}
\end{figure}

\section{Research problems}

In this area, some research problems remain. 

First of all, after a thorough search of relevant literature, it seems that no researchers have yet applied the character-level convolutional network to the Chinese character. Only some of the NLP tasks are based on pinyin encoding. Pinyin input is one of the most popular forms of text input in the Chinese language\cite{Chen:2000:NSA:1075218.1075249}. This method represents the Chinese characters in an alphabetic way according to its pronunciation (See Figure 1.1). Previous researchers such as Mathew used pinyin format dataset as input to detect spam in mobile text message\cite{mathew2011intelligent} and Liu, similarly, using pinyin format dataset for feature selection\cite{liu2010short}, proved that pinyin format dataset could be utilised as an efficient method to solve the NLP problems. Unlike the English language, there is no gap between the characters in the Chinese language. However, the word segmentation still plays a significant role in understanding the sentence meaning. 

Secondly, how can the Chinese corpus gain benefit from the ConvNets when not only the pinyin encoding but also the Chinese character lack language root? As we observed, the previous datasets mainly involve English corpus, language roots such as prefix and suffix are indicated to contribute to ConvNets \'s ability to solve the typo problem in NLP. 

Thirdly, the information compression among pinyin format, Chinese characters, and English may lead to different performances. For example, in the Chinese language, there are only 406 syllable combinations can be found in pinyin format representation among more than 6,000 commonly used Chinese characters, which means that some information is compressed during the transforming\cite{Chen:2000:NSA:1075218.1075249}. Another example is between English and Chinese, Twitter and Weibo, two popular online social networks, have a 140-word limit for each posting. However, since two different language systems, Roman alphabet and Chinese character alphabet, are mainly used respectively on Twitter and Weibo. Twitter can only provide titles and short web links while Weibo can afford more detailed information. Moreover, since more than 100 possibilities of word or character choice are available for the same pronunciation, it remains unclear whether the ConvNets can gain a representation from pinyin dataset in text classification task. 

The tasks are yet to be solved due to the lack of large-scale Chinese character dataset, and we believe that by concentrating on the following three parts, we can solve these tasks. Firstly, we will compare our proposed model with previous models on the pinyin format dataset. In this particular task, pinyin encoding on text classification task, we will find out the most important factors according to different experiments such as various depth and choice of alphabet. Secondly, the neural network model often requires a large-scale dataset so that the model can better extract features for classifying. To solve the problem of missing Chinese character dataset, an entirely new Chinese character dataset and its corresponding pinyin encoding dataset will be constructed. Finally, we will evaluate our models on these two datasets to find out the better solution.

\section{Contribution}

Firstly, this thesis applies the character-level convolutional neural network to Chinese character dataset, which is rarely researched in this NLP area. The result shows that Chinese character dataset has generated a better result than its corresponding pinyin encoding. Secondly, we have reached the state-of-the-art in this specific task among all the other narrow character-level convolutional neural network. Besides, we are the first one who constructed a large-scale Chinese character dataset. Moreover, we have extended the pinyin format that reached millions level compared with the previous one.

\section{Structure}
In this chapter, we have provided some background information concerning natural language processing, deep learning models, and Chinese language. Also, we have outlined the research questions and fundamental information about our model. Following this, Chapter 2 discusses and compares related work of text classification. Then in Chapter 3, we describe the architecture of our model. Next, in Chapter 4, we show the details of experimental results and hyper-parameter settings during the process. We also introduce our latest constructed datasets. After this, some observation is discussed. Finally, in Chapter 5, we conclude the work that has been done so far and forecast the future directions.

\chapter{Related work}

\begin{table}[h]
\centering
\begin{tabular}{llll}
\hline
Authors & Model          & Pros                      & Cons                  \\ \hline
Harris  & Bag of Words   & Easy to understand        & High-dimensional sparse          \\
William & N-gram   & Accurate                  & High-dimensional sparse                \\
Armand  & Bag of Tricks  & Fast and simple           & Comparable result     \\
Hiyori  & Graph Boosting & Graph representation      & Complicated structure \\
Zhang   & CNN            & Construct large-scale dataset     & Too many parameters   \\
Xiao    & CNN + BiLSTM   & Combined different models & Complicated structure \\
Conneau & CNN            & Very Deep layers          & Time-consuming       
\end{tabular}
\caption{The comparison between different related work models.}
\end{table}

In terms of text classification, various researchers employ different algorithms. These approaches follow the same scheme, which is feature extraction followed by classification. Traditional models include SVM \cite{joachims1998text}, Naive Bayes \cite{mccallum1998comparison}, Bag of Words \cite{harris1954distributional}, N-gram\cite{cavnar1994n}, and their TF-IDF \cite{sparck1972statistical} version. In previous researches, these algorithms have been evaluated, and most of them provide a competitive result.

\section{Traditional models}
\subsection{Bag of Words}
The early references of Bag of Words concept can be found in ``Distributional structure" by Harris. This model uses the counts of the most common words that appear in the training set as the feature\cite{harris1954distributional}. This model can classify the topics with the help of these keywords. For example, the three keywords ``Dow" ``Jones" ``Indexes" may have a much more frequent appearance in the articles of the stock topic than sports topic. However, since some of the words appear in each of these topics, this may influence the result. That is why the TF-IDF version Bag of Words add one extra feature, the inverse document frequency, to diminish this influences. This model is also based on the word-level, which means many words that share the same word stem but not the same count. In some way, by using the stemming technique, we can avoid this problem, yet not all the words containing same word stem have approximative meaning, which may lead to another problem. The result in Zhang\'s paper\cite{zhang2015character} shows that Bag of Words model and its TF-IDF version can achieve great performances in most of the tasks.

\subsection{N-gram}
The N-gram model in text classification task can be seen as an extension of the Bag of Word model\cite{cavnar1994n}. An N-gram model is commonly used in the language model. Unlike the Bag of Word model above, the N-gram model uses the most frequent N-continuous word combinations selected from the dataset as the features. For instance, the model would calculate the appearance of word combination ``Dow Jones Indexes" in all of the topics before applying the predefined class that ranks the most. TF-IDF version also adds the inverse document frequency to avoid the common words problem. This model is widely used in NLP area because of its trait, simplicity, and scalability. Zhang\'s result shows that N-gram model achieved excellent performances in the small dataset, especially the TF-IDF version, ranking the first in three of the dataset.

\subsection{Bag of tricks}
Bag of tricks is a straightforward and efficient approach for text classification. This algorithm can train the model in no more than ten minutes on one billion words, and classify a large number of datasets among millions of classes within a minute\cite{joulin2016bag}. It is one of the greatest models so far as a traditional model in this area. In one hand, this model can be trained in a fast speed. On the other hand, the result is quite close compared with the state-of-the-art that using character-level ConvNets.

\subsection{Graph representation classification}
Yoshikawa\cite{yoshikawa2016fast} proposed a fast training method for text classification which is based on graph classification. This method treats the input text as a graph, and the graph structure can better represent the text structure. The result shows that graph representation can exploit rich structural information of texts, and this is the key to improving their accuracy.

\section{Neural Network models}
There is also a large amount of research using deep learning methods to solve the text classification problem. 

\subsection{Recrusive Neural Network}
A recursive neural network often comes with a parser. In Socher's\cite{socher2013recursive} work, a parse tree is being used in the feature extraction stage. However, most of the dataset will not come with a parser, which means this kind of model is not general enough. As we observed, no related models are being released these two years.

\subsection{Recurrent Neural Network}
A recurrent neural network is like a particular recursive neural network. This model brings in the data sequentially, mostly from left to right, sometimes it may be bidirectional. Liu \cite{liu2012sentiment} solved the sentiment analytics task using this model. An embedding layer followed by a recurrent layer is used to extract feature and then fed into the classification layer.

\subsection{Convolutional Neural Network}
Research has also shown that ConvNets is effective for NLP tasks \cite{collobert2011natural}\cite{collobert2008unified}, and the convolutional filter can be utilised in the feature extraction stages.

Kim was one of the earliest researchers who used convolutional neural networks (ConvNets) for sentence classification\cite{kim2014convolutional}.

In this paper, Kim\cite{kim2014convolutional} proposed a word-level shallow neural network with one convolutional layer using multiple widths and filters followed by a max-pooling layer over time. The fully-connected layer with drop-out layer can then combine features and send to the output layer. The word vectors in this model were initialised using the publicly available word2vec, which was trained on 100 billion words from Google News\cite{mikolov2013efficient}. The comparison between several variants and traditional models applied on six datasets are reported in this paper, they are movie reviews with one sentence per review \cite{pang2005seeing}, TREC question dataset \cite{li2002learning}, a dataset for classifying the sentence whether subjective or objective\cite{pang2005seeing}, customer reviews of various products \cite{hu2004mining}, opinion polarity detection subtask of the MPQA dataset \cite{wiebe2005annotating}, in particular, Stanford Sentiment Treebank \cite{socher2013recursive}. The classes of those datasets are between two to six, and the dataset size is from 3,775 to 11,855. Kim selected the stochastic gradient descent (SGD) and Ada-delta update rule\cite{zeiler2012adadelta} for his model.

The paper shows that the unsupervised pre-training of word vectors is an important part of word-level ConvNets for NLP.  Also, the temporal k-max pooling layer can capture and provide much more capacity because of the multiple filter widths and feature maps. Finally, dropout layer proved to be a good regulariser that adds performance. However, there are still some ways to explore. Only one convolutional layer was constructed in this ConvNets, while the trend in computer vision where significant improvements have been reported using much deeper networks, 19 layers\cite{simonyan2014very}, or even up to 152 layers\cite{he2015deep}. Also, due to the small size of the dataset, the word-level ConvNets for NLP is yet to prove.

Zhang was the first one who proposed an entirely character-level convolutional networks for text classification\cite{zhang2015character}. 

There are two different ConvNets illustrated in this paper and the difference between them is the feature map size. Both of their depth are eight, consists of six convolutional layers and two fully-connected layers. The pooling layers follow the convolutional layers.  Convolutional kernels of size seven are used in the first two layers, and the rest of the four layers\' kernel size are three. There are also two dropout layers between the three fully-connected layer to regularise the loss. The input of this model is a sequence of encoding vectors, which is done by `applied an alphabet of size n for the input documents, and then quantise each character using one-hot encoding\cite{zhang2015character}'. However, Zhang did not use any pre-trained method such as word2vec\cite{mikolov2013efficient} for the input word vectors in the model. By Contrast, Zhang used data augmentation techniques to enhance their performances by an English thesaurus. Also, they distinguish the upper-case and lower-case letters. However, the results show that worse result when such distinction is made. Moreover, Zhang constructed eight large-scale datasets to fill the vacancy in this NLP task. The eight large-scale datasets are Ag\'s News, Sogou News corpus, DBPedia, Yelp Reviews, Yahoo Answers, and Amazon Reviews. Their size is from 120,000 to 3,600,000, and the classes are between two and fourteen. A comparison was made among traditional models such as SVM\cite{joachims1998text}, Naive Bayes \cite{mccallum1998comparison}, Bag of Words, N-grams\cite{cavnar1994n}, and their TF-IDF version\cite{sparck1972statistical}, also among deep learning models such as word-level ConvNets and long-short-term memory model(LSTM)\cite{hochreiter1997long}. The result shows that Bag-of-Means were the worst models among all 22 different model\'s setting, and their testing error is the highest among all the dataset. N-gram model gained the lowest error rate in Yelp Review Polarity, and its TF-IDF version reached the best result in AG\'s News, Sogou News, DBPedia. Different settings of the ConvNets attained the lowest error rate in the rest of four datasets.

The paper shows that character-level ConvNets is an efficient and potential method due to their performance mentioned above. As we can observe in the result, when the dataset size comes large, the ConvNets models can do better than traditional models. Also, ConvNets work well for user-generated data, which means the ConvNets may be suitable in real-world scenarios. Then, for large-scale datasets, the distinction between the selection of upper letters and lower letters may lead to a worse result. However, the result is varied indicates that no single model works for all of the tasks or datasets.

Conneau\cite{conneau2016very} was the first one who implemented a very deep convolutional architecture which is up to 29 convolutional layers (9, 17, 29, 49 respectively) and applied to sentence classification. 

There is a look-up table creating vectorial representations fed into the model, with a convolutional layer behind. Then a stack of temporal ``Convolutional blocks" which are a sequence of two convolutional layers, each one followed by a temporal batch-normalization layer and a ReLU activation function. Different depths of the overall architecture are obtained by varying the number of convolutional blocks in between the pooling layers. The k-max pooling layer is followed to obtained the most important features of the stack of convolutional blocks\cite{conneau2016very}. Finally, the fully-connected layers output the classification result. In this model, Conneau did not use `Thesaurus data augmentation' or any other preprocessing, except lower-case the texts. The datasets in this paper are the same corpora of Zhang and Xiao \cite{xiao2016efficient}, and the best results of them are the baseline in this article. The results show that the deep architecture works well when the depth increased, and the improvements are significant especially on large data sets compared with Zhang\'s convolutional models.

Most of the previous application of shallow ConvNets to NLP tasks combining different filter size together. They indicate that the convolution layers can extract N-gram features over tokens. However, in this work, Conneau \cite{conneau2016very} created an architecture which used many layers of small convolutional kernel, which is size three. Compared with Zhang’s ConvNets architecture, Conneau\cite{conneau2016very} found that better not to use dropout layer with the fully-connected layers, but only temporal batch normalization\cite{ioffe2015batch} after convolutional layers. At the same time, Conneau evaluates the impact of shortcut connections by increasing the number of convolution to 49 layers. As described in He\cite{he2015deep}, the gain in accuracy due to the increase of the depth is limited when using standard ConvNets. To overcome the degradation problem, He introduced ResNet model that allow gradients to flow more easily in the network. Moreover, as Conneau observed, they found improvement results when the network has 49 layers. However, they did not reach state-of-the-art results under this setting. 

\subsection{Convolutional Neural Network and Recurrent Neural Network}
There are also some researches that combine both ConvNets and Recurrent Neural Network\cite{xiao2016efficient}\cite{tang2015document}.

Xiao\cite{xiao2016efficient} combined both convolutional network and recurrent network to extract the features and applied to sentence classification. 

The following model contains embedding layer, convolutional layers, recurrent layers and the classification layers which are one, three, one, and one layers respectively. The convolutional network with up to five layers is used to extract hierarchical representations of features which serve as input for an LSTM. The word vectors were not pre-trained, while there is an embedding layer to transform the ``one-hot vectors" into a sequence of dense vectors. The eight datasets in this paper are the same as Zhang, and Xiao, show that it is possible the use a much smaller model to achieve the same level performance when a recurrent layer is added on top of the convolution layers. There are five datasets that Xiao reached a better result than Zhang, which is AG’s News, Sogou, DBPedia, Yelp Review full and Yahoo Answers. The rest of the result are closed. However, the parameters in Xiao’s model are much less than Zhang\'s model. 

Compared with the character-level convolutional neural network model, this model achieved comparable performances for all the eight datasets. By reducing the number of convolutional layers and fully-connected layers, this model is with significantly fewer parameters, up to 50 times less, which means that they generalised better when the training size is limited. Also, this paper shows that the recurrent layer can capture long-term dependencies to solve the problem that convolutional layers usually require many layers due to the locality of the convolution and pooling. Moreover, the model achieves a better result when the number of classes increases compared with ConvNets models. It is mainly because the less pooling layer in hybrid models can preserve more detailed and complete information. Finally, there is an optimal level of local features to be fed into the recurrent layer, because Xiao noticed that the model accuracy does not always increase with the depth of convolutional layers\cite{xiao2016efficient}.

In summary, all of the related work models are based on Roman alphabet corpus. In this paper, we will describe our character-level convolutional neural network model applied to both Chinese character and its pinyin format and report the result on the latest large-scale dataset.
\chapter{Proposed solution}

\begin{figure}[t]
\centering 
\fbox{\includegraphics[height = 17cm]{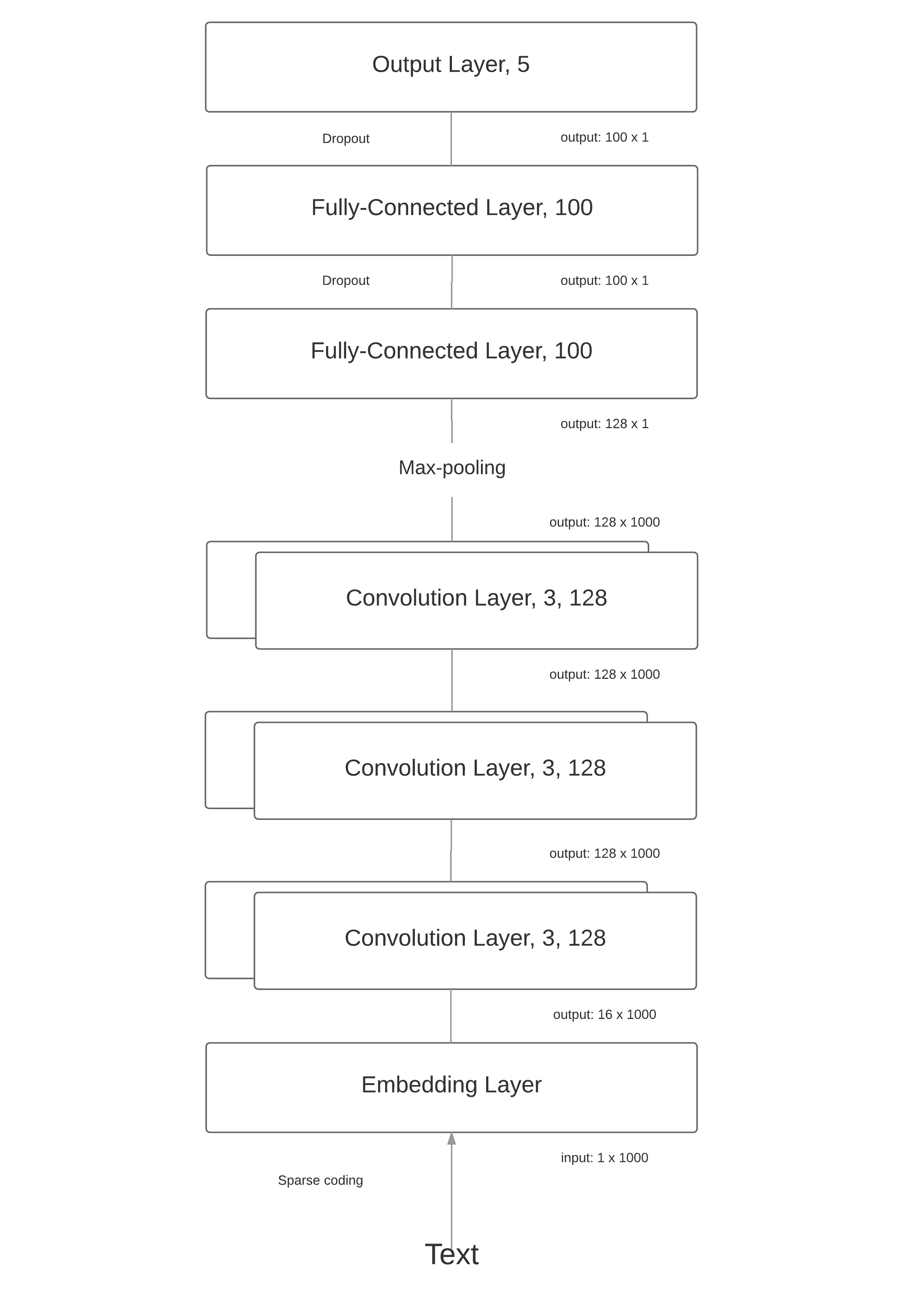}}
\caption{The architecture of proposed model. The number `3, 128' in the convolutional layer represents the kernel size and feature maps number respectively. The `100' in the fully-connected layer represent the output dimension. The `5' in the output layer indicates that there are five pre-defined labels.}
\end{figure}

In this chapter, we present the model architecture in detail. There are four components in our model, which are data preprocessed, embedding layer, convolutional layers and fully-connected layers.

\section{Data preprocessed}
Our model begins with a data preprocessed stage. This stage is used for transforming the original characters to encoded characters. There are two types of the encoding method (See Figure 3.1) which are one-hot encoding and one of m encoding (Sparse coding). For the one-hot encoding, each word in the sentence is represented as a one-hot vector. The i-th symbol in this vector is set to one if it is the i-th element, while the rest of the symbol are remain zero\cite{zhang2015character}. However, we use the one of m encoding in our model. For the input corpus, we firstly construct an alphabet size equal to \emph{S}, then used this dictionary to quantise each character, and the characters which are not in the alphabet will be replaced by a blank. Also, we set the maximum length of each sequence to \emph{L}, and the exceeding part will be ignored, and the missing part will be replaced by zero, using zero-padding. Therefore, we can get a dense \emph{S} sized vector with solid size \emph{L}.

\begin{figure}[h]
\centering 
\fbox{\includegraphics[height = 6cm]{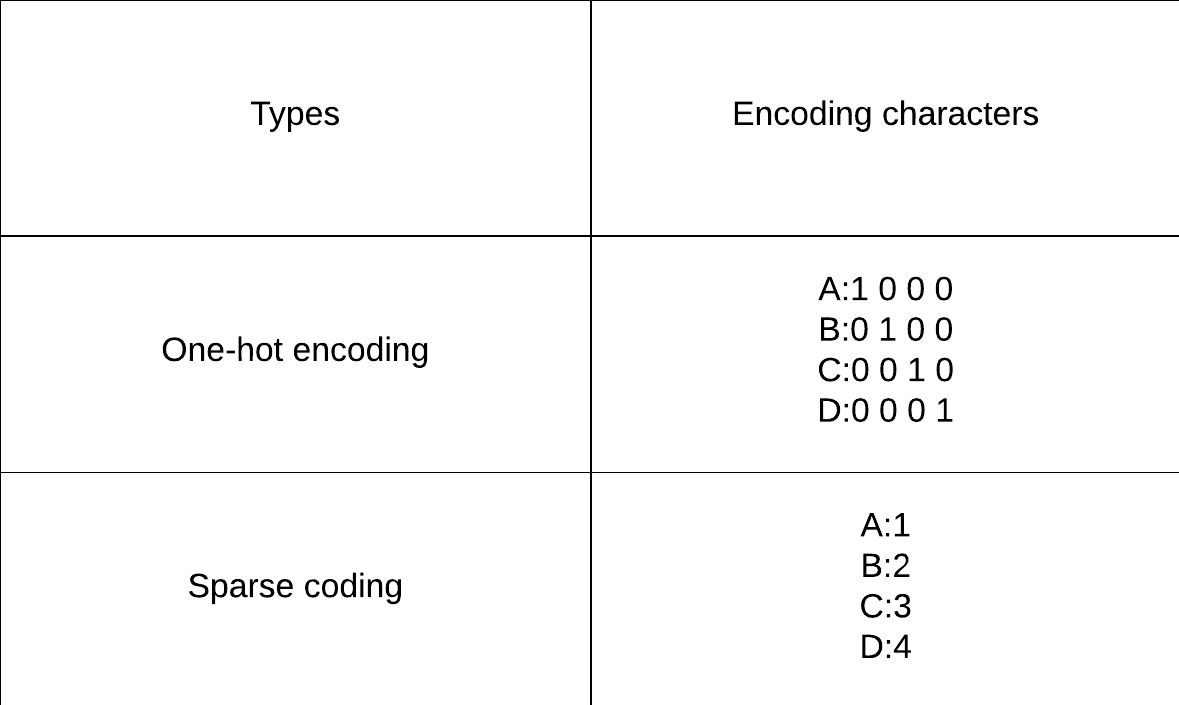}}
\caption{Two kinds of encoding. The input characters are a, b, c, and d, while the encoding matrix can represent the characters respectively.}
\end{figure}

According to previous researches, the quantisation order will be reversed. Usually, when we receive the information, the latest content will leave a deeper impression compared with the early one. According to this assumption, by reversing the encoding order, the most recent content can help the fully-connected layer to gain a better result\cite{zhang2015character}.

Due to the differences between English and pinyin, the alphabet for pinyin input are different. In previous researches\cite{zhang2015character}\cite{conneau2016very}, the alphabet size are 71 and 72 characters respectively. However, according to our experiment result, some trait of pinyin encodings such as no uppercase letter in the alphabet and not all 26 Roman letters included in the pinyin alphabet, therefore, we picked 42 characters to construct our dictionary. The smaller alphabet size efficiently decreases the parameters and improves the performances. For the Chinese character input, the dictionary is much larger than the pinyin input, reached to 6,653 characters, which are the words that appears in the dataset. The different size of dictionary between pinyin and Chinese character can help us understand how the dimension influences the result.

\section{Embedding Layer}

As we can see from the Fig 3.2, embedding Layer accepts a two-dimensional tensor of size \emph{S} * \emph{L} which is the encoded character sequence. Usually, the embedding layer is used for decreasing the dimension of the input tensor. Also, the zero-padding help to transform the input tensor into a fixed size. However, as we have finished these two process in the previous stages, we can naturally treat the embedding layer as a look-up table. After the processed in embedding layer, the output shape can be treated as an image of size \emph{S} * \emph{L}, while the \emph{S} is like the `RGB' dimension in computer vision. By converting the input character into the one-dimension vector, the ConvNets can then extract the feature by the convolutional kernel.

\section{Convolutional Layer}

In convolutional layers, we apply up to three 1D-Convolution layers which have kernel size equal to three and feature map equal to 128. The operation of convolution is widely used in signal and image processing. For the 1D-Convolution we used, there are two signals which are text vector and kernel. After the process, the convolutional operation created a third signal which is the output. The text vector \emph{f} is the output from the embedding layer, and the \emph{g} is the kernel. For the text vector \emph{f}, its length is \emph{n} equal to 1,000 in our setting (for Chinese character is 300) while kernel \emph{g} has length \emph{m} which is 3. Here is the definition of the operation between \emph{f} and \emph{g} (The formula is also refer to Cornell University CS1114 courses, section6) :

$$(f * g)(i) = \sum_{j=1}^{m} g(j) * f(i - j + m/2)\cite{zhang2015character}$$

We can imagine the calculation is the kernel slides from the very beginning to the end (including the zero-padding part), so that the 1D-convolution can extract features from the text input. Every kernel will be represented as a specific feature. 

Also, each layer is initialised by \emph{\textbf{henormal}}\cite{he2015delving}, and the border mode is \emph{\textbf{same}}, which means the length remains 1,000 during this stages. In previous researches such as N-gram language model, researchers combine a different kind of N-gram such as bigram and trigram as the features, to extract multiple combinations of words from the dataset. The results show this setting can decrease the perplexity in the language model. However, latest work suppose that we can use deeper model and unify kernel size to extract features efficiently. Encouraged by Conneau’s work\cite{conneau2016very}, we only used the kernel size equal to three, so that during this period, the layers can automatically best combine these different ``trigram" features in various layers.

We used the \emph{\textbf{ReLU}}\cite{nair2010rectified} as our nonlinear activation function, which is widely employed in recent researches. Unlike the `sigmoid’ function, this activation function can better handle the gradient vanishing problem. Also, the threshold in \emph{\textbf{ReLU}} can better simulate the brain mechanism of human. L2 regulation is being used in all these layers because it is quite efficient for solving the overfitting problem. The output of the convolutional layer is a tensor of size 128 * 1000. They are the hierarchical representations of the input text. The convolutional layers can automatically extract N-gram features from the padded text, and these features can represent the hidden long short term relation in the text. In computer vision area, convolutional kernels are able to construct the pattern from the very beginning units, such as pixel, line, and shape. Meanwhile, the structure in NLP is similar, including character, word, and sentence. The similar properties in these two areas make the ConvNets interpretable.

Finally, the max-pooling layer followed by the convolutional layer is necessary. There are varies types of pooling layers such as max-pooling layer and average pooling layer\cite{conneau2016very}. The pooling layer can select the most important features from the output of 1D-convolution. Also, it can diminish the parameters to accelerate the training speed. We chose the temporal max-pooling with kernel size equal to the feature maps number, which means only the most important feature remains in this stage. At last, by using the flatten function, these features will be sent to the fully-connected layer with size 128 * 1 as a 1D tensor.

\section{Fully-Connected Layer}

The fully-connected layer also knows as the dense layer. At this stage, all the resulting features that selected from the max-pooling layer are combining. As we mention earlier, the max-pooling layer selects the k-most feature from each convolutional kernel. The fully-connected layer can combine most of the useful assemble and then construct a hierarchical representation for the final stage, the output layer. 

The output layer used `softmax' as the nonlinear activation function, and there is five neurones because of the number of the target classes. Unlike the current state of the art, Conneau did not use any dropout layers between the fully-connected layers. For a very deep architecture model, the batch normalization layer may be a better choice. However, because our model is not that deep, we still apply the dropout layer and set the dropout rate to `0.1'.

\chapter{Results and Discussion}

In this chapter, we will present our results and findings from the following two tasks. The first task is about the factors that may influence the ConvNets when the dataset is pinyin format. Compared with the previous models, we can find out the best setting for pinyin format. Following this, we evaluated our model on pinyin format dataset with task 1 setting and Chinese character dataset in task 2. The detail information is as follow.

\section{The factors that influenced the ConvNets when applied to pinyin format dataset}

\subsection{Task description}

In this task, we validated our models on one of the eight datasets in Zhang\cite{zhang2015character} ‘s research. This dataset aiming to solve the news categorisation problem is widely used in different researches. The dataset was collected from Sogou\cite{wang2008automatic}, and the encoding is pinyin format. By comparing our models with previous researches on the same dataset, we prove that our model can achieve state-of-the-art result with fewer parameters and faster training speed among all the narrow ConvNets. 

\subsection{Dataset Description}

In computer vision area, there are many large datasets used for image classification or object detection, such as ImageNet\cite{ILSVRC15}, CIFAR\cite{krizhevsky2009learning} and their size are millions level with more than 1,000 classes. In text classification area, the Sogou pinyin dataset is one of the eight large-scale datasets that Zhang constructed. All the dataset are character level, and the pinyin one contains five classes, with all equal size.

\subsection{Model setting}

Here are the settings that have been used in our experiments. By comparing among different hyper parameters, these settings were found to be best for this specific task.

The dictionary for the ConvNets needs to adjust to the certain context although pinyin format encoding and English corpus are both based on the Roman alphabet. In previous researches, researchers need to distinguish between upper-case and lower-case letters\cite{zhang2015character}\cite{xiao2016efficient}\cite{conneau2016very}, which means the dimension of the dataset is at least fifty-two due to the Roman alphabet size. The worse result has been found when such distinction is made. Zhang explained that the differences between letter cases might affect the semantics, and that may lead to a regularisation problem\cite{zhang2015character}. However, in pinyin format encoding dataset, there is no distinction between the upper-case and lower-case letters. Furthermore, we only added four basic punctuations into the dictionary to lower the dimension. Figure 4.1 shows the dictionary:

\begin{figure}[h]
\centering 
\fbox{\includegraphics[width = 12cm]{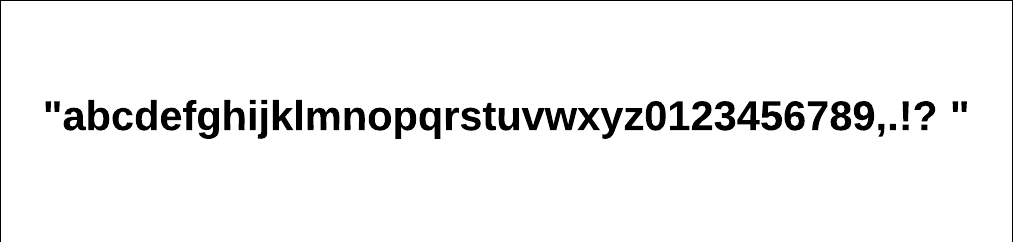}}
\caption{The dictionary of pinyin format encoding dataset (including a blank character).}
\end{figure}

As we observed from the original news article statistics, most of the content are no more than one thousand characters. Therefore, during the data preprocessed stage, all of the input text padded to a fixed size equal to 1000, and the following embedding layer convert them into a dense vector with dimension 16. Every convolutional layer applies to the same setting. The convolutional kernel size is three, and there is two zero-padding part located two sides. With this two zero-padding part, the length among the convolutional layers remains the same (length = 1000), which is useful for stacked layer structure. We initialised the convolutional layers using Gaussian initialisation scaled \cite{he2015delving}. We did not use any pre-trained method because our model is entirely based on character level, while the word-level are usually required the pre-trained method to avoid the local optimum problem. We set the dropout rate to `0.1' between the fully-connected layers. The dropout layers are proved to be useful to avoid the overfitting problem so that we can gain a similar result in both training set and test set. The batch size is 128, which means every time the model updated the parameters after 128 data trained via the back propagation process. Training performed with optimiser `Adam'\cite{kingma2014adam}, and the loss functions is `categorical cross-entropy'. Compared with other optimisers such as SGD, the `Adam' can converge faster and guide to a better result. Also, because of the model is a multi-class classification problem, we used the `categorical cross-entropy' and `softmax' activation in our output layer. All the rest of the hyper-parameters are configured to default in the model. The implementation is done via Tensorflow and Keras on a single NVIDIA GeForce GTX TITAN X GPU. 

\subsection{Result and Discussion}

Our ConvNets reached the state-of-the-art in narrow convolutional network. According to our comparison between different models, the proposed model reached state-of-the-art when the convolutional layers are restricted to seven layers. Meanwhile, the parameters of our model are up to 190 times less than the other models. The detailed information about result and discussion are listed below.

\textbf{Choice of dictionary is important}

\begin{figure}[t]
\centering 
\fbox{\includegraphics[width = 14cm]{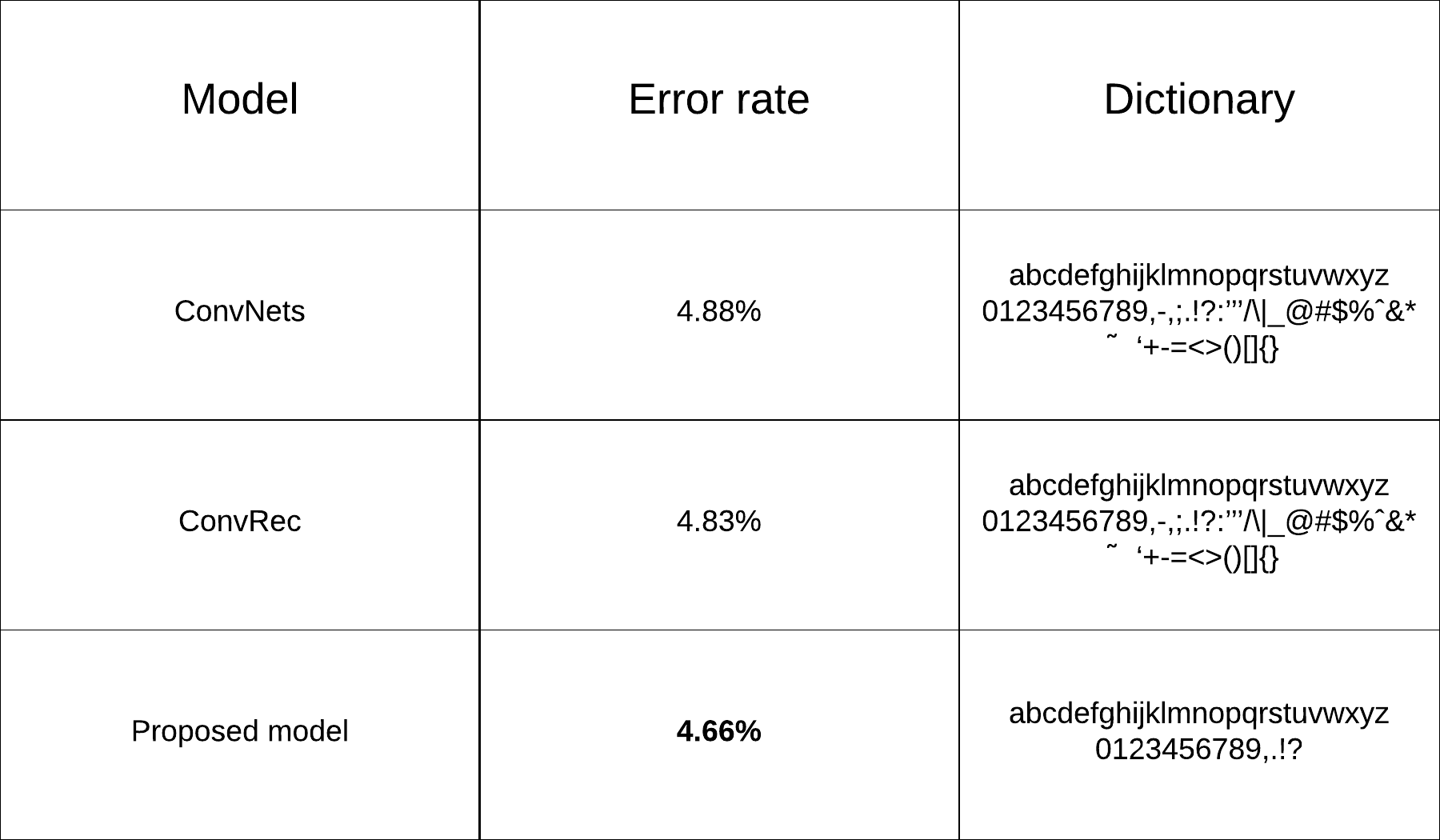}}
\caption{The comparison between different model\'s dictionary.}
\end{figure}

The result in Figure 4.2 shows that the chosen of the dictionary is one of the most important factors in character-level ConvNets. With the help of an appropriate dictionary, we achieved state-of-the-art in narrow convolutional network. Two reasons may lead to this result.

Firstly, in previous researches, the dictionary is not suitable for pinyin format encoding dataset. In English corpus, 26 Roman letters are being used in all of the English articles. However, there are still differences between pinyin alphabet and English alphabet. For instance, the letter `v' are not included in pinyin alphabet. Also, there are no upper letters in pinyin format alphabet. These different rules may lead to various ways to construct the dictionary, and it will influence the performance.

Secondly, the dictionary may affect the replacement operation in pre-processed stage. During the pre-processed stage, we need to use the `regular expressions' to replace the word to a blank character if they are not in the dictionary. With an adjusted dictionary, the useless characters will be replaced by a blank character. This stage can be seen as we removed the noise from an original song which can help to extract the features from the dataset.

Finally, the appropriate dictionary can significantly decrease the parameters and improve the speed. Fewer parameters mean that we can train the model not only on some specifically graphic card such as TITAN X with 12 GB memory, also on some other graphic card with 4 or 8 GB memory.

\textbf{ConvNets need stacked layers with proper hyper-parameters}

The result in Table 4.1 shows that with fine-tuning parameters, we can reach state-of-the-art with fewer parameters. The best model for the proposed one demonstrates that we can use deeper layers with smaller feature maps rather than a single layer with a large feature maps size. It is because the trait of ConvNets is to extract the features in a partial space. By combining the hierarchical representations in the fully-connected layer, we can let the model choose appropriate features automatically. For instance, three convolutional layers with feature maps number equal to 128, the parameters are about two times large than one convolution layers with feature maps number equals to 350. After all, the stacked one provides a better result and requires fewer parameters.

\begin{table}[]
\centering
\begin{tabular}{lllll}
\hline
Model             & Parameters    & Error rate      & Network structure & Feature maps \\ \hline
Original ConvNets & 27000k        & 4.88\%          & CNN6-FC2          & 1024         \\
ConvRec           & 400k          & 4.83\%          & CNN3-Rec1         & 128          \\ \hline
Proposed ConvNets & 90k           & 7.69\%          & CNN1-FC1          & 350          \\
Proposed ConvNets & 84k           & 5.84\%          & CNN2-FC1          & 128          \\
Proposed ConvNets & 133k & 4.72\% & CNN3-FC1          & 128          \\
Proposed ConvNets & \textbf{140k} & \textbf{4.66\%} & CNN3-FC2          & 128          \\ \hline
\end{tabular}
\caption{The comparison between previous models and various proposed models. The table including parameters, error rate, network structure, and the feature maps hyper-parameters. The best model\'s error rate is in bold.}
\end{table}

The result in Table 4.1 also shows that the depth of our character-level ConvNets influences the performances. The output pf convolutional layers are the most important part of this model because they can gain the hierarchical representations of the context, and this is the key for following layer to classify the classes.

\section{The comparison between Chinese character and its corresponding pinyin format dataset}

\subsection{Task description}

In this task, we validated our models on two large-scale datasets. The first one is pinyin format encoding dataset, and another one is Chinese character dataset. These datasets are collected by Sogou\cite{wang2008automatic} and then reallocate and transformed for this paper. By comparing their performances, we can prove that character-level ConvNets works better on Chinese character dataset. Also, we will discuss the theory behind this model.

\subsection{Dataset Description}

In text classification area, there are not any large-scale Chinese character dataset exist. Therefore, we decided to create a new dataset with Chinese character and its corresponding pinyin version. The size of the dataset is up to 1,150,000 by using the data augmentation. 

\begin{table}[t]
\centering
\begin{tabular}{lllll}
\hline
Dataset           & Classes & Training Size & Test Size & Overall Size \\ \hline
Chinese character & 5       & 490,717       & 86,597    & 577,314      \\
Pinyin            & 5       & 490,717       & 86,597    & 577,314      \\
Pinyin*           & 5       & 981,434       & 173,194   & 1,154,628    \\ \hline
\end{tabular}
\caption{The comparison between different datasets, including the Chinese character and pinyin format. The star indicates that dataset expand by data augmentation.}
\end{table}

We combine the news corpus SogouCA and SogouCS from Sogou Lab\cite{wang2008automatic}, which is more than 3 million news articles and at least twenty categories. We labeled the dataset by their domain names, which is part of their URL link. Also, our dataset only contains five categories, which are sports, finance, alternating, automobile, and technology, because of not all the categories have enough data. These datasets are range from 50,000 to 20,000, which are the top five classes in the original news article. If the length of any news content is less than 20 words, the corresponding data will be removed. After all the pre-processing work, there are five classes in the dataset, with about 1200k training set and 200k test set. In Zhang\'s dataset, the size of each class is equal, while our latest dataset has different size in each category so that we can observe whether the ConvNets can extract the feature correctly.

The data augmentation is useful for deep learning models to enhance their performances. This technique is widely used in computer vision \cite{he2015deep} and speech recognition \cite{abdel2012applying}, to increase the size of the dataset by transforming the signals or rotating the image. In our dataset, the original corpus is the Chinese character. We used the Python library \emph{pypinyin} and \emph{jieba} library to transform the dataset from original Chinese character to pinyin format encoding. Then, we used two different pinyin format which is provided by the Python library \emph{pypinyin}, to enlarge the dataset, and the experiment shows it helps in some ways.

\subsection{Model setting}

The dictionary of Chinese character is distinct from the pinyin format. There are more than six thousand common used Chinese characters which means the dimension of the dictionary is quite large. In this task, we used all of the Chinese character that appears in both training set and test set to construct the dictionary, and the dictionary size is equal to 6,653. Usually, we need more than one characters in pinyin format to represent a Chinese character (See Figure 4.4), so that we changed the input length to 250 according to the statistics of the training set. The remain setting of the model is the same as the model of pinyin format dataset.

\subsection{Result and Discussion}

We compared the result of Chinese corpus and pinyin format encoding. To the best of our knowledge, this is the first time that character-level convolutional network applied to Chinese characters. We discussed our potential finding below.

\begin{figure}[hb]
\centering 
\fbox{\includegraphics[width = 15cm]{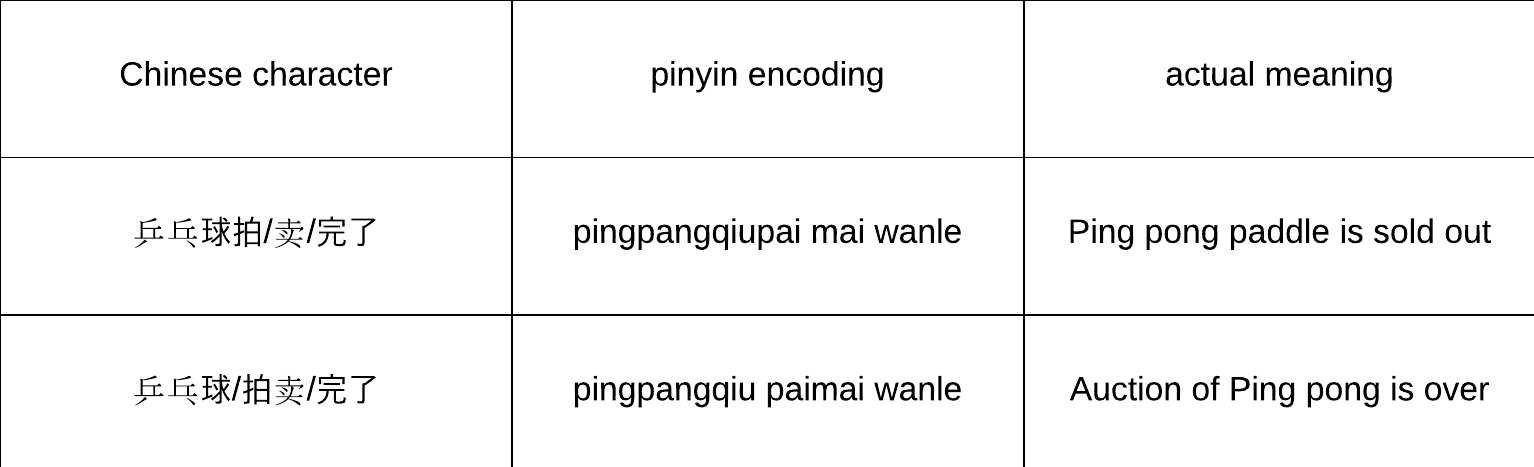}}
\caption{A misunderstanding example indicates the importance of word segmentation in Chinese language.}
\end{figure}

\textbf{The character-level ConvNets solved the word segmentation problem in Chinese}

\begin{table}[t]
\centering
\begin{tabular}{lllll}
\hline
Model         & Accuracy        & Parameters & Layout  & Dataset encoding  \\
Bag of Words* & 6.55\%          &            &         & pinyin A          \\
N-gram*       & 2.81\%          &            &         & pinyin A          \\\hline
ConvNets      & \textbf{5.47\%} & 127k       & CNN1-FC1 & Chinese character \\
ConvNets      & 6.15\%          & 177k       & CNN2-FC2 & Chinese character \\
ConvNets      & 6.88\%          & 236k       & CNN3-FC2 & Chinese character \\
ConvNets      & 10.98\%         & 20k        & CNN1-FC1 & pinyin A          \\
ConvNets      & 8.5\%           & 143k       & CNN3-FC2 & pinyin A          \\
ConvNets      & 11.15\%         & 34k        & CNN1-FC1 & pinyin B          \\
ConvNets      & 8.61\%          & 128k       & CNN3-FC2 & pinyin B          \\
ConvNets*     & 10.94\%         & 35k        & CNN1-FC1 & augmentation      \\
ConvNets*     & 8.87\%          & 129k       & CNN3-FC2 & augmentation     
\end{tabular}
\caption{The comparison between different setting and encoding dataset. The results show that Chinese character works better than pinyin format. The result of Bag of Words and N-gram came from Zhang\cite{zhang2015character}, which are the references of this task. The ConvNets with star means the dataset are expanded via data augmentation.}
\end{table}

The Table 4.3 shows that Character-level ConvNets works well on both pinyin format encoding and Chinese character corpus. The error rate of pinyin encoding is between 8.41\% and 8.87\%, while the Chinese character corpus reached 5.47\%. The results are impressive, which shows the character-level ConvNets can extract the features from the Chinese character dataset efficiently. We believe that this is the first work that shows character-level ConvNets can be applied to Chinese character in text classification task.

We assume that character-level ConvNets works well on Chinese character due to the solved of word segmentation problem. The word segmentation plays a major role in the Chinese language because different segmentation indicates varied meaning. The Figure 4.3 shows that for the same sentence, different word segmentation may lead to a different meaning. In character-level ConvNets, the unit is character level, which means convolutional layers extract the character combination as a feature and ignore the segmentation among the words. In previous researches, character-level ConvNets has already proved that they can better handle the misspelling problem and typo problem in the Roman alphabet. The researchers assume that the convolutional kernel can ignore the differences between two words shared same language roots such as suffix and prefix\cite{zhang2015character}. However, the pinyin format dataset is based on pronunciation. The dataset can not gain benefit from the language root, also need to suffer from the information compressed during the transforming process, and that is why the Chinese character dataset performs better than the pinyin one in our model.

\begin{figure}[t]
\centering 
\fbox{\includegraphics[width = 13.5cm]{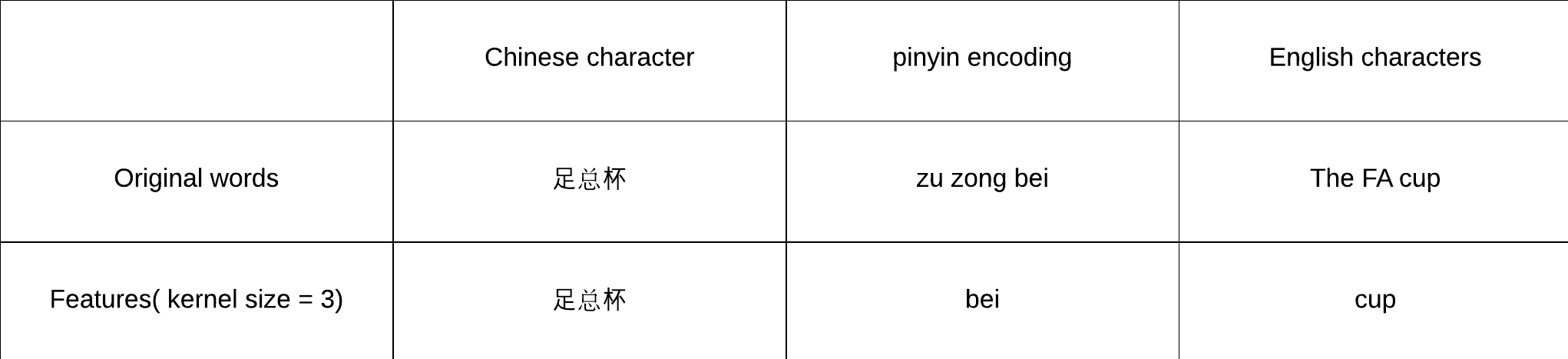}}
\caption{For the same length, Chinese characters contain more information.}
\end{figure}

\textbf{Chinese character dataset suffers from overfitting problem}

The Chinese character can hold more information than pinyin format. As Figure 4.4 shows, a Chinese character requires several Roman characters to represent. That means under the same convolutional kernel size, feature maps setting; the Chinese character dataset can storage more information than the pinyin encoding one. Moreover, this is the main reason as stacked layers increased, the error rate raised from 5.47\% to 6.88\% due to the overfitting problem. We assume that there are potential to decrease the error rate with fine-tuning parameters. Also, the character-level ConvNets may perform better when the Chinese character dataset become larger.

\textbf{The pinyin format dataset increases the information in some way}

As we mentioned in the introduction chapter, with more than six thousand common used Chinese characters, there are only four hundred and six syllable combinations, which means there is some information compressed during the process. However, the results show that pinyin format dataset presents competitive result, and this leads to a question, how does it happen?

Here are the reasons we have found in the analytics. Firstly, the word segmentation helps. Unlike the western language, there is no gap between each Chinese words, so during the transforming process, the \emph{pypinyin} python library required another python library \emph{jieba} to assist the word segmentation. That means the blank character between each word can provide a more accurate meaning that helps both human and computer to understand. Secondly, the words represent different meanings according to the context, sometimes even native speaker can misunderstand the actual meaning. However, the pinyin format encoding can handle this problem with the help of pronunciation annotation so that the human can now precisely get the meaning via its pronunciation, so does the computer.

\textbf{To some extent, data augmentation helps to improve the performances}

In our dataset, we use two different ways to construct the pinyin encoding format dataset, and the combined dataset size reached about 1.15 million. The error rate seems increased with this combined dataset. However, to some extent, it helps improve the performances. As the Table 4.2 shows, the dataset is combined with two different encoding. If the convolutional neural network extract one specific feature that comes from pinyin encoding type A, this feature can not be used in the test set with pinyin encoding type B, even they represent the same meaning in this context. Moreover, we assume this will cause a significant decrease in the accuracy. Nevertheless, the result remains almost the same, which means the data augmentation do improve the results in some way. In the future, we can refer to computer vision area, to extend the dataset via some appropriate transformations, such as synonyms and paraphrasing, so that the data augmentation technique can efficiently control generalisation error because of the model can better handle the overfitting problem.

\chapter{Conclusion}

We have presented a character-level convolutional neural network that is focusing on solving text classification problem in Chinese corpus. This model has been evaluated on brand new datasets which are consists of Chinese characters and its corresponding pinyin format, and the sizes are up to 600 thousand and 1.15 million respectively. The evaluation between Chinese characters and its corresponding pinyin format dataset shows that character-level ConvNets can works better with Chinese character dataset. We believe that the main reasons are due to the success of solving the word segmentation problem and the information storage capacity of Chinese corpus. To the best of our knowledge, this is the first time that the character-level ConvNets applied to Chinese corpus to solve the text classification task. Also, the validation on Zhang\'s dataset\cite{zhang2015character} show that proper dictionary and hyper-parameters play a major role in pinyin format text classification task.

In the future, we believe that the similar ideas can apply to the other NLP tasks such as reading comprehension, machine translation. With the help of character-level ConvNets, we may fulfil the full potential of Chinese language in NLP. Also, it will be interesting to apply a pre-trained method to character-level ConvNets when applied to Chinese character. Because the character in Chinese are meaningful as a word, while the character in English is meaningless. Finally, we hope that there will be more large-scaled Chinese dataset contains more classes in NLP area, in order to increase the difficulty of the task.
\addcontentsline{toc}{chapter}{Appendices}

\appendix
\chapter{User Manual}

The source code and dataset are all available from Github link:

https://github.com/koalaGreener/Character-level-Convolutional-Network-for-Text-Classification-Applied-to-Chinese-Corpus

\section{Requirements}

Python2 / Python3

Keras

Tensorflow

(Run this model under CUDA will significantly increase the training speed, Nvidia graphic card required.)

\section{Components}

This repository contains the following components:

Models: Different model settings for the various dataset. There are four different folders, which are 2008\_models, Chinese\_character, pinyin\_formatA, pinyin\_formatB. 2008\_models represents the task 1 \' model which is applied to Zhang\'s dataset. The other models are all for task 2, and the datasets are constructed for this paper. 

Dataset: The dataset for proposed model. The 2008\_dataset is constructed by Zhang in his paper, while the 2012\_dataset is constructed for this paper, including pinyin format and Chinese character dataset.

Data Preprocessing: Some source code for data preprocessing, and it can be used to reproduce the dataset using Sogou news articles.

\section{Example Usage}

1.Install the Keras and switched the backend to Tensorflow.

2.Download the dataset from Google drive and put both the training set and test set into the data folder.

3.Import the ipynb files into the IPython Notebook and start to train the models.

\section{Licence}

MIT



\addcontentsline{toc}{chapter}{Bibliography}

\bibliography{Main}

\end{document}